\documentclass[pre,twocolumn,showpacs,preprintnumbers,10pt]{revtex4-1}

\usepackage{amsfonts,amssymb,amsmath,amsthm,latexsym,times}
\usepackage{graphicx, epsfig, color}
\usepackage[english]{babel}
\usepackage{float}
\usepackage{enumitem}
\usepackage{siunitx}

\mathchardef\mhyphen="2D

\begin{document}

\title{Optimal model-free prediction from multivariate time series
}

\author{Jakob Runge$^{1,2)}$, Reik V. Donner$^{1)}$, and J\"urgen Kurths$^{1,2,3,4)}$}
\affiliation{ $^{1)}$ Potsdam Institute for Climate Impact Research, P.O. Box 60 12 03, 14412 Potsdam, Germany\\
            $^{2)}$ Department of Physics, Humboldt University, Newtonstr. 15, 12489 Berlin, Germany\\
$^{3)}$ Institute for Complex Systems and Mathematical Biology, University of Aberdeen, Aberdeen AB24 3UE, United Kingdom \\
$^{4)}$ Department of Control Theory, Nizhny Novgorod State University, Gagarin Avenue 23, 606950 Nizhny Novgorod, Russia
        }

\date{\today}
\begin{abstract}
Forecasting a time series from multivariate predictors constitutes a challenging problem, especially using model-free approaches. Most techniques, such as nearest-neighbor prediction, quickly suffer from the curse of dimensionality and overfitting for more than a few predictors which has limited their application mostly to the univariate case. 
Therefore, selection strategies are needed that harness the available information as efficiently as possible. 
Since often the right combination of predictors matters, ideally all subsets of possible predictors should be tested for their predictive power, but the exponentially growing number of combinations makes such an approach computationally prohibitive.
Here a prediction scheme that overcomes this strong limitation is introduced utilizing a causal pre-selection step which drastically reduces the number of possible predictors to the most predictive set of causal drivers making a globally optimal search scheme tractable. 
The information-theoretic optimality is derived and practical selection criteria are discussed. As demonstrated for multivariate nonlinear stochastic delay processes, the optimal scheme can even be less computationally expensive than commonly used sub-optimal schemes like forward selection.
The method suggests a general framework to apply the optimal model-free approach to select variables and subsequently fit a model to further improve a prediction or learn statistical dependencies. The performance of this framework is illustrated on a climatological index of El Ni\~no Southern Oscillation.
\end{abstract}


\maketitle

\section{Introduction} \label{sec:introduction}
Predicting the future behavior of complex systems from measured time series constitutes a major goal in many fields of science.
Traditionally, this problem has been mainly addressed in terms of model-based approaches, often using linear models \cite{Brockwell2002}. 
As an alternative, since the late 1980s also model-free predictions have been developed using nearest neighbors in state space \cite{Sidorowich1987,Abarbanel1990,Giona1991,Alparslan1998,Pompe2001,Ragwitz2002} or neural networks \cite{Eisenstein1995,Szpiro1997,Small2002}. In the nearest-neighbor technique, states similar to the present state are searched for in the past of a time series $Y$ and a future value $Y_{t+h}$ at a prediction step $h$ is forecasted by simply averaging the $Y$ values $h$-steps ahead of the nearby past states or using local-linear models \cite{Sidorowich1987}. 
Model-free predictions have so far been mostly univariate where states are usually reconstructed from embedding a single time series using Taken`s theorem \cite{Takens1981,Sauer1991,Ragwitz2002}. However, the intertwined nature of complex systems calls for multivariate approaches taking into account more available information. 
Now the problem is that the curse of dimensionality makes useful nearest-neighbor predictions almost impossible for more than a few predictors \cite{Hastie2009}, especially if many of these predictors carry redundant information.

From an information-theoretic perspective, the minimal set of variables that maximizes the (multivariate) mutual information \cite{Cover2006} with a target variable is most predictive \cite{Groth2001,Pompe2001}. Minimality is required to avoid the curse of dimensionality. It is important to note that this set of variables can be different from those with \emph{individually} large mutual information with the target variable. Indeed, sometimes the right combination of predictors matters. For example, if $Y$ is driven multiplicatively by $X\cdot Z$, the mutual information of each of these predictors with $Y$ can be very low and only the mutual information of the combined set $(X, Z)$ with $Y$ is very high. In general, such \emph{synergetic} sets can only be detected by searching through all subsets of variables. However, the number of possible combinations for taking into account more variables and larger time lags grows exponentially making such a search strategy prohibitive due to computational constraints.

Therefore, simple search strategies such as ranking the predictors by their mutual information with a target variable or the \emph{CMI-forward selection} using conditional mutual information (CMI) have been proposed recently \cite{Kugiumtzis2013}. 
Here we demonstrate that such approaches can fail already in simple cases where one cannot avoid to test different subsets of predictors. However, we information-theoretically prove that the search can be restricted to \emph{causal} drivers. To obtain these drivers, we exploit a recently developed model-free algorithm to consistently reconstruct causal drivers from multivariate time series that keeps the estimation dimension as low as possible with typically low computational complexity \cite{Runge2012prl}. The much smaller set of causal drivers then allows for a globally optimizing search strategy which is optimal also for synergetic cases. In this contribution, we additionally provide a practical criterion for selecting the optimal size of the subset of predictors which compares well even with computationally expensive cross-validation approaches. In numerical experiments we found that our optimal scheme yields much improved predictions and often even runs faster than forward selection.

Our method suggests a general framework not only for prediction, but also for general statistical inference problems for datasets (not only time series) where the underlying mechanisms are poorly understood: Firstly, the optimal model-free approach can be applied for selecting not only causal, but also possibly synergetic driving variables and, secondly, these variables can be used to fit a model to learn the functional form of the dependencies on these causal predictors. This approach combines the advantage of a model-free approach to detect relevant variables with the advantage of model-based methods to efficiently harness these variables to further improve predictions or understand mechanisms. Our framework is illustrated on a sea-surface temperature based index of the El Ni\~no Southern Oscillation (ENSO) in the tropical Pacific.

This paper is organized as follows. After deriving the optimality of causal predictors in Sect.~\ref{sec:optimal_prediction}, we discuss common approaches for information-theoretic variable selection for predictions in Sect.~\ref{sec:common_prediction_scheme}. The optimal scheme is explained in Sect.~\ref{sec:optimal_prediction_scheme} including the causal pre-selection algorithm and selection criteria. Section~\ref{sec:computational_complexity} discusses the computational complexity of the different schemes. In Sect.~\ref{sec:model_example} we compare our scheme with other approaches in a model example. Extensive numerical experiments on multivariate nonlinear stochastic delay processes are conducted in Sect.~\ref{sec:numerical_experiments}. In Sect.~\ref{sec:optimal_plus_model} we analyze the combination of the model-free selection with a model-based prediction scheme which is applied in Sect.~\ref{sec:enso_prediction} to predict future values of the considered ENSO index.

\section{Optimal prediction} \label{sec:optimal_prediction}
The discrete-time evolution of a subprocess $Y$ of a multivariate $N$-dimensional stochastic process $\mathbf{X}$ can be described by 
\begin{align}  \label{eq:evolution}
Y_{t+1} &= f\left(\mathcal{P}_{Y_{t+1}},\, \eta^Y_{t+1}\right)\,,
\end{align} 
with some function $f(\cdot)$, where $\mathcal{P}_{Y_{t+1}}\subset \mathbf{X}^-_{t+1}=(\mathbf{X}_{t},\,\mathbf{X}_{t-1},\,\ldots)$ are the deterministically driving variables including the past of $Y$ at possibly different time lags and $\eta^Y_{t+1}$ represents stochastic noise driving $Y$. The uncertainty about the outcome of $Y_{t+1}$ can be quantified by the Shannon entropy \cite{Cover2006} which decomposes into 
\begin{align} \label{eq:entropy_decomp}
H(Y_{t+1}) &=  I(Y_{t+1}\,;\,\mathbf{X}_{t+1}^-) + H(Y_{t+1}\,|\,\mathbf{X}_{t+1}^-)\,,
\end{align}
where the latter term is the \emph{source entropy} \cite{Shannon1948,Pompe2011}. This conditional entropy quantifies the minimum level of uncertainty that cannot be predicted even if the whole past (and present) $\mathbf{X}_{t+1}^-$ is known. If the dependency of $f$ on the noise term $\eta^Y_{t+1}$ is additive, the source entropy equals the entropy of the noise: $H(Y_{t+1}\,|\,\mathbf{X}_{t+1}^-)=H(\eta^Y_{t+1})$.
The infinite-dimensional multivariate mutual information (MMI) $I(Y_{t+1}\,;\,\mathbf{X}_{t+1}^-)$, on the other hand, quantifies the predictable part by measuring by how much the uncertainty about the outcome of $Y_{t+1}$ can be maximally reduced if $\mathbf{X}_{t+1}^-$ was perfectly measured. 

In practice, a prediction using the entire set $\mathbf{X}_{t+1}^-$ (truncated at some maximal lag) would severely suffer from the curse of dimensionality and \emph{overfitting} \cite{Hastie2009}, which means that many variables do not actually carry useful information, but merely fit the noise in the time series, and the prediction -- trained on a learning set -- would perform poorly on a test set.
The goal is, thus, to select a small subset of predictors from $\mathbf{X}_{t+1}^-$ that carries a maximum of information about $Y_{t+1}$ and still generalizes well on new data. However, for an $N$-variate process $\mathbf{X}$ truncated to a maximum delay $\tau_{\max}$ the number of possible subsets grows exponentially in $N$ and $\tau_{\max}$. 

To avoid this combinatorial explosion, simple search strategies such as ranking the individual predictors by their mutual information (MI-scheme) with the target variable can be used which, however, is prone to include redundant variables that do not improve a prediction. Forward selection, a more advanced technique, iteratively determines predictors based on how much information they contain \emph{additionally} to the already chosen variables using conditional mutual information (CMI-scheme) \cite{Kugiumtzis2013} leading to a polynomial computational complexity. These strategies will be discussed in Sect.~\ref{sec:common_prediction_scheme}. 
But forward selection is not a globally optimal strategy, one reason being that it might select variables that are not deterministically driving $Y_{t+1}$, the other that it fails to detect synergetic cases as demonstrated in our model example in Sect.~\ref{sec:model_example}. 

The unknown deterministic drivers $\mathcal{P}_{Y_{t+1}}$ in Eq.~(\ref{eq:evolution}) are, however, key to arrive at optimal predictors as can be shown by decomposing the MMI in Eq.~(\ref{eq:entropy_decomp}) using the chain rule \cite{Cover2006}
\begin{align} \label{eq:mmi_decomp}
&I(Y_{t+1};\mathbf{X}_{t+1}^-) = I\left(Y_{t+1}\,;\,\mathcal{P}_{Y_{t+1}} \right)+\nonumber\\
& \phantom{I(Y_{t+1};\mathbf{X}_{t+1}^-) =} + \underbrace{I\left(Y_{t+1}\,;\,\mathbf{X}_{t+1}^-\setminus \mathcal{P}_{Y_{t+1}} \,|\,\mathcal{P}_{Y_{t+1}}\right)}_{= 0}\,, 
\end{align}
where $\setminus$ denotes the set subtraction. The second term is zero for processes satisfying the \emph{Markov property} which states that $Y_{t+1}$ is independent of the remaining past given its \emph{causal parents} $\mathcal{P}_{Y_{t+1}}$, a term that originates from the theory of \emph{graphical models} \cite{lauritzen1996graphical}. For multivariate time series these are called \emph{time series graphs} \cite{Eichler2011,Runge2012prl,Runge2014a}.
This proves that, theoretically, all information is already contained in the causal parents. Adding a variable from $\mathbf{X}_{t+1}^-$ would not increase the information, but removing a parent from $\mathcal{P}_{Y_{t+1}}$ would decrease the MMI in Eq.~(\ref{eq:mmi_decomp}). The causal parents can be efficiently estimated by the algorithm described in Sect.~\ref{sec:optimal_algo}. The Markovity rests on the assumption that the noise term $\eta^Y_{t+1}$ driving $Y$ is independent of the noise terms driving the other variables. While this assumption is not strictly fulfilled in many real world systems, it often at least approximately holds. This assumption is also not as crucial for the prediction task as it is for the causal inference problem.

However, for finite time series, some predictors in $\mathcal{P}_{Y_{t+1}}$, even though causal, could only be weakly driving and lead to overfitting since they do not generalize well on new data. It is, therefore, crucial to optimize the selection of a minimal subset of causal parents. Since the set of causal parents $\mathcal{P}_{Y_{t+1}}$ is much smaller than $\mathbf{X}_{t+1}^-$, this can now be done using a global optimization strategy. In Sect.~\ref{sec:selection_criteria} we present such a strategy to select the \emph{optimal subset $\mathcal{P}^{(p)}_{Y_{t+1}}\subseteq\mathcal{P}_{Y_{t+1}}$ of causal predictors}.

\begin{figure}[!t]
\begin{center}
\includegraphics[width=1.\columnwidth]{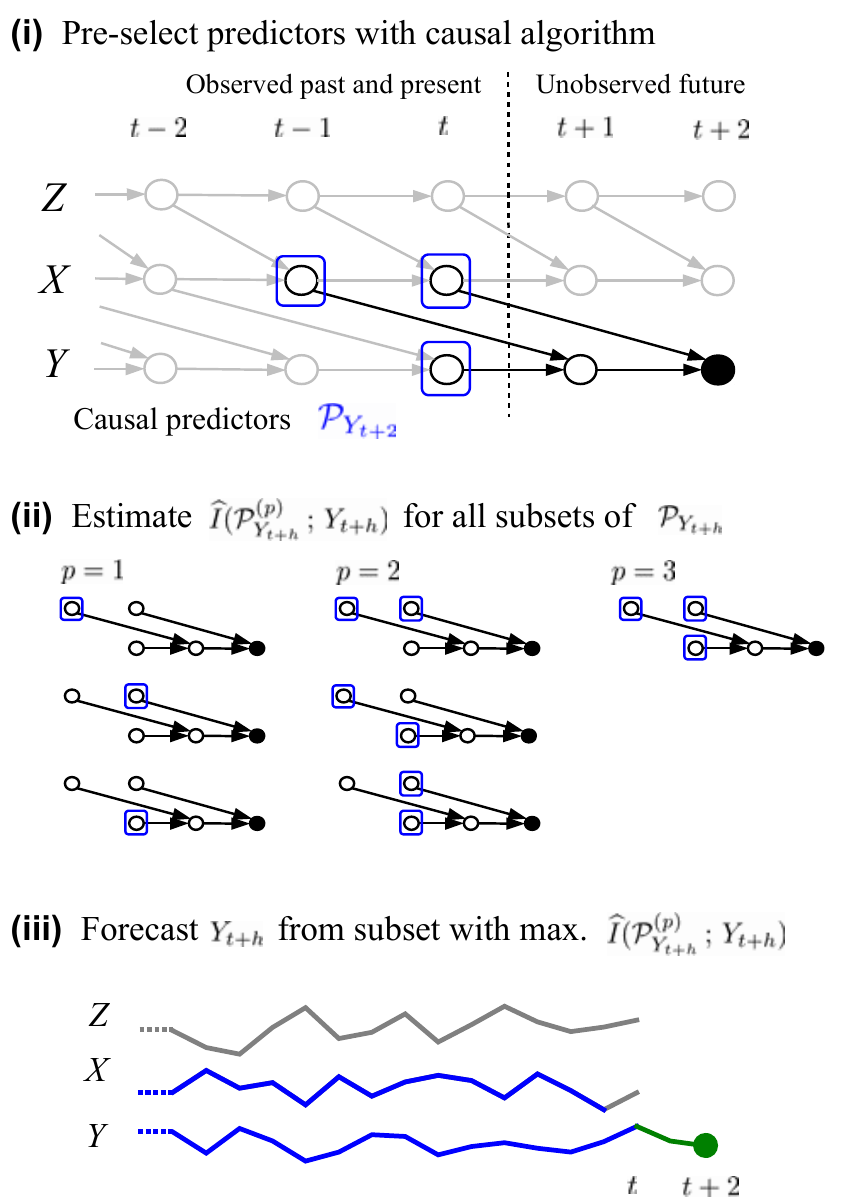}
\end{center}
\caption[]{(Color online)
Optimal prediction scheme. 
(i) Causal pre-selection in an example time series graph (see text) for predicting $Y_{t+2}$. Even though $Z$ could have a high mutual information with $Y$, its influence is only indirect through the parents $X_t$ and $Y_{t+1}$ of $Y_{t+2}$. Among these the latter already lies in the unobserved future, but part of its information can still be recovered by measuring $Y_{t}$ and $X_{t-1}$ which share information along the paths marked with black arrows. These variables form the causal predictors $\mathcal{P}_{Y_{t+h}}$ (blue boxes) for $h=2$, which can be found by determining the Markov set. A suitable algorithm for this task will be discussed in Sect.~\ref{sec:optimal_algo}.  All paths from nodes further in the past to $Y_{t+2}$ have to pass through this set.
(ii) Selection of optimal subset from causal predictors. For all numbers of predictors $p$, the multivariate mutual information $\widehat{I}(\mathcal{P}^{(p)}_{Y_{t+h}}\,;\,Y_{t+h})$ of all possible subsets is estimated.
(iii) The optimal predictors are those where the estimated MMI takes its maximum or can be obtained from cross-validation. For example, if the optimal predictors were $(Y_t,\, X_{t-1})$, only time series samples (blue shaded) of these predictors are used with a nearest-neighbor (Sec. IV C) or model-based (Sec. VIII) prediction scheme to forecast the future value t+h of the time series of Y.
}
\label{fig:prediction_scheme}
\end{figure}

For predictions $h>1$ steps into the future, the set of causal predictors $\mathcal{P}_{Y_{t+h}}$ for $Y_{t+h}$ is not identical with the parents anymore as for $h=1$ because predictors can only be chosen among the observed variables $\mathbf{X}^-_{t+1}=(\mathbf{X}_{t},\,\mathbf{X}_{t-1},\,\ldots)$ prior to $t+1$. Still the same algorithm as for the causal parents can be used to obtain the set of variables that separates $Y_{t+h}$ from $\mathbf{X}^-_{t+1}\setminus \mathcal{P}_{Y_{t+h}}$ in the time series graph. In Fig.~\ref{fig:prediction_scheme}(i) an example of such a graph is given. As defined in \cite{Runge2012prl}, each node in that graph represents a subprocess of a multivariate discrete time process $\mathbf{X}$ at a certain time $t$. Nodes are connected by a directed link if they are not independent conditionally on the past of the whole process, which implies a lag-specific \emph{Granger causality} with respect to $\mathbf{X}$ \cite{Eichler2011}. Using these causal predictors, the only uncertainty left comes from the source entropy of $Y_{t+h}$ and the entropy from the unobserved ancestors of $Y_{t+h}$ between $t{+}1$ and $t{+}h{-}1$ (see Fig.~\ref{fig:prediction_scheme}(i)). 

In the following sections we discuss and numerically compare the four prediction schemes mentioned above: (1)~MI selection, (2)~CMI-forward selection, (3)~CMI-forward selection of only causal predictors, and (4)~our optimal scheme.

\section{MI and CMI prediction schemes} \label{sec:common_prediction_scheme}

In the first prediction scheme, MI-selection, the respective MI of each variable in $\mathbf{X}_{t+1}^-$ up to a maximum lag $\tau_{\max}$ with the target variable $Y_{t+h}$ is estimated. Then the predictors $X^{(\cdot)}\in \mathbf{X}_{t+1}^-$ are ranked by their MI: $I(X^{(1)}; Y_{t+h})>I(X^{(2)}; Y_{t+h})>I(X^{(3)}; Y_{t+h})>\ldots$. To determine the best number $p$ of the ranked predictors that should be used, one can either apply a heuristic criterion or \emph{cross-validation} \cite{Hastie2009}. In the model experiments in Sects.~\ref{sec:model_example} and \ref{sec:numerical_experiments} we evaluate both approaches, employing the heuristic criterion that the MI of the ranked predictor $X^{(p)}$ should be at least a fraction $\lambda$ of the MI of the previous predictor $X^{(p-1)}$, i.e., $I(X^{(p)}; Y_{t+h})>\lambda ~ I(X^{(p-1)}; Y_{t+h})$ with $\lambda\in [0,1)$. The last of the ranked predictors that satisfies this criterion determines the selected number $\widehat{p}$ of predictors. 
This scheme has the drawback that two or more predictors with high MI with the target variable might contain highly redundant information. Then the inclusion of redundant predictors leads to overfitting which will be discussed in Sect.~\ref{sec:model_example}.

The second prediction scheme, CMI-forward selection, overcomes this drawback by excluding information already contained in the previous predictors \cite{Kugiumtzis2013}: First the MIs $I(X_{t-\tau};Y_{t+h})$ for all $X_{t-\tau}\in \mathbf{X}^-_{t+1}$ are estimated. The first predictor $X^{(1)}$ is the one that maximizes the MI with the target variable (i.e., the same one as in the MI-selection scheme). The next predictor $X^{(2)}$, however, is chosen according to the maximal CMI $I(X_{t-\tau};Y_{t+h}|X^{(1)})$ among all remaining predictors, the third predictor is the maximum CMI conditional on the two previously selected predictors, etc. 
In each step $p$, the CMI gives the gain to the MMI if this predictor is included because
\begin{align} \label{eq:cmi_decomp}
& I((X^{(1)},\ldots,X^{(p)});Y_{t+h}) \nonumber \\
&= \underbrace{I((X^{(1)},\ldots,X^{(p-1)});Y_{t+h})}_{\text{MMI without}~X^{(p)}} \nonumber\\
&\phantom{=}+  \underbrace{I(X^{(p)};Y_{t+h}|(X^{(1)},\ldots,X^{(p-1)}))}_{\text{gain due to}~X^{(p)}}.
\end{align}
Here the heuristic criterion \cite{Kugiumtzis2013} is to select as the best $\widehat{p}$ the last $p$ with at least a gain of 
\begin{align} \label{eq:heuristic_optimal_prediction}
&I(X^{(p)};Y_{t+h}|(X^{(1)},\ldots,X^{(p-1)})) \nonumber \\
&>\lambda ~ I((X^{(1)},\ldots,X^{(p-1)});Y_{t+h})\,,
\end{align}
where $\lambda\in [0,1)$ as before. In \cite{Kugiumtzis2013} also an adaptive choice of $\widehat{p}$ using a shuffle test is discussed.
This scheme has been proposed to infer causal drivers in \cite{Kugiumtzis2013}. However, it can be shown to fail for this task already in simple cases which will be shown in Sect.~\ref{sec:model_example}. 

Rather than with a heuristic criterion, at the cost of additional computation time, the best number $p$ of predictors can also be chosen by cross-validation. Here we use an $m$-fold cross-validation where the available observed set of time indices $\mathcal{T}=\{1,\ldots,T\}$ is partitioned into $m$ complementary segments. For each validation round, a fold $m$ is retained as the \emph{validation set} $\mathcal{T}_m$ on which the prediction performance is evaluated. The nearest-neighbors are searched for in the complementary set $\mathcal{T} \setminus\mathcal{T}_m$ from which also the prediction estimate is generated. Then the number $\widehat{p}$ of predictors where the average prediction error across all $m$ folds is minimal is chosen.

\section{Optimal prediction scheme} \label{sec:optimal_prediction_scheme}
For the prediction of $Y_{t+h}$ given the multivariate time series $\mathbf{X}$, our proposed optimal prediction scheme consists of the following steps (Fig.~\ref{fig:prediction_scheme}): (i)~Estimate the causal predictors $\mathcal{P}_{Y_{t+h}}\subset \mathbf{X}^-_{t+1}$ using the causal algorithm described in the next Sect.~\ref{sec:optimal_algo}, (ii)~check all subsets (except the empty one) and select the $p$ causal predictors $\mathcal{P}^{(p)}_{Y_{t+h}} \subseteq \mathcal{P}_{Y_{t+h}}$ with the maximal \emph{estimate} of the MMI $\widehat{I}(\mathcal{P}^{(p)}_{Y_{t+h}}\,;\,Y_{t+h})$ with the target variable as the optimal ones (Sect.~\ref{sec:selection_criteria}), (iii)~use these predictors to forecast the target variable with nearest-neighbor prediction (Sect.~\ref{sec:nnp}).
In the following, we explain the causal pre-selection algorithm and discuss criteria for selecting the optimal subset. While here the actual prediction is conducted using a nearest-neighbor scheme \cite{Sidorowich1987}, in Sect.~\ref{sec:optimal_plus_model} we will also discuss how a model-based prediction based on the inferred optimal predictors can further improve a forecast.

\subsection{Causal pre-selection algorithm} \label{sec:optimal_algo}
The causal pre-selection algorithm is a modification of the algorithm introduced in \cite{Runge2012prl}, which is based on the PC algorithm \cite{Spirtes1991,Spirtes2000} (named after its inventors Peter Spirtes and Clark Glymour). The main idea is to iteratively unveil the causal predictors by testing for independence between pairs of nodes in the time series graph conditional on a subset of the remaining nodes. Since these conditions are efficiently chosen, the dimension stays as low as possible in every iteration step. This important feature helps to alleviate the curse of dimensionality in estimating CMIs \cite{Runge2012prl} affecting the computation time as well as the reliability of conditional independence tests. 
Under some mild assumptions discussed below, the algorithm yields a consistent estimate of $\mathcal{P}_{Y_{t+h}}$.
Instead of the commonly used binning estimators where the curse of dimensionality is especially severe, here we utilize an advanced nearest-neighbor estimator \cite{Frenzel2007} that is most suitable for variables taking on a continuous range of values. This estimator has as a free parameter the number of nearest neighbors $k$ which determines the size of hyper-cubes around each (high-dimensional) sample point. Small values of $k$ lead to a lower estimation bias but higher variance and vice versa. Therefore, we choose different values in the algorithm ($k^{\rm algo}$) and the subsequent selection schemes ($k^{\rm CMI/MMI}$). Note that for an estimation from (multivariate) time series stationarity is required.

The algorithm starts with no \emph{a priori} knowledge about the drivers and iteratively learns the set of predictors of $Y$:
First, estimate unconditional dependencies $I(X_{t-\tau}; Y_{t+h})$ and initialize the preliminary predictors $\mathcal{P}_{Y_{t+h}}=\{X_{t-\tau}\in \mathbf{X}^-_{t+1} : I(X_{t-\tau}; Y_{t+h})  > 0 \}$. This set contains also non-causal predictors which are now iteratively removed by testing whether the dependence between $Y_{t+h}$ and each $X_{t-\tau}\in\mathcal{P}_{Y_{t+h}}$ conditioned on the incrementally increased subset $\mathcal{P}^{n,i}_{Y_{t+h}} \subseteq \mathcal{P}_{Y_{t+h}}$ vanishes:
\begin{enumerate}

    \item[($n.$)] Iterate $n$ over increasing number of conditions, starting with some $n_0>0$:
    
        \begin{enumerate}
             
             \item[($n.i$)] Iterate $i$ through all combinations
                          of picking $n$ nodes
                     from  $\mathcal{P}_{Y_{t+h}}$ to define the conditions $\mathcal{P}^{n,i}_{Y_{t+h}}$ in this step, and
                         estimate
                         $I(X_{t-\tau}; Y_{t+h}\,|\,\mathcal{P}^{n,i}_{Y_{t+h}})$ for all 
                         $X_{t-\tau} \in \mathcal{P}_{Y_{t+h}}$.
                      After each step the nodes $X_{t-\tau}$ with 
                        $I(X_{t-\tau}; Y_{t+h}\,|\,\mathcal{P}^{n,i}_{Y_{t+h}}) = 0$ are removed from
                     $\mathcal{P}_{Y_{t+h}}$ 
                         and the iteration over $i$ stops if all possible 
                         combinations have been tested. 
                         (In the implementation we limit the number $n_i$ of combinations and check relevant combinations first, see \cite{Runge2012prl,Runge2014d} for details.)
        \end{enumerate}
        
    If the cardinality $|\mathcal{P}_{Y_{t+h}}| \leq n$, the algorithm converges,
    else, increase $n$ by one and iterate again. 
    (In the implementation we limit the dimensionality up to some $n_{\max}$. If the initial number of conditions is $n_0>1$ to speed up the algorithm, also previously skipped combinations with $n<|\mathcal{P}_{Y_{t+h}}|$ need to be checked before convergence can be assessed.)

\end{enumerate}
The main assumptions underlying the identification of the conditional independence structure with the PC algorithm are the \emph{Causal Markov Condition}, i.e., Markovity of the process of any finite order, and \emph{Faithfulness}, which guarantees that the graph entails all conditional independence relations true for the underlying process and can be violated only in certain rather pathological cases \cite{Spirtes2000}.
If these assumptions are fulfilled, the causal algorithm is universally consistent, implying that the algorithm will converge to the true causal predictors with probability 1 for infinite sample size \cite{Spirtes2000}. On the other hand, the CMI forward selection algorithm proposed for causal inference in \cite{Kugiumtzis2013} is not consistent since it yields non-causal drivers already in simple cases which will be analyzed in Sect.~\ref{sec:model_example}. But the CMI forward selection scheme can be `repaired' by a further backward elimination step \cite{Sun2014}.

Practically, the causal algorithm involves conditional independence tests for $I(X_{t-\tau}; Y_{t+h}\,|\,\mathcal{P}^{n,i}_{Y_{t+h}}) = 0$. 
In \cite{Runge2012prl} a shuffle test is proposed for testing whether $I(X_{t-\tau}; Y_{t+h}\,|\,\mathcal{P}^{n,i}_{Y_{t+h}})>0$: An ensemble of $M$ values of $I(X^*_{t-\tau}; Y_{t+h}\,|\,\mathcal{P}^{n,i}_{Y_{t+h}})$ is generated where $X^*_{t-\tau}$ is a shuffled sample of $X_{t-\tau}$, i.e., with the indices permuted. Then the CMI values are sorted and for a test at a given $\alpha$-level, the $\alpha M$-th value is taken as a significance threshold. In \cite{Runge2012prl} a numerical study on the detection and false positive rates of the algorithm are given. 
The shuffle test comes at the additional cost, that for each conditional independence test $M$ surrogates of CMI have to be estimated. An alternative is to apply a fixed threshold $I^*$, which has the drawback that it does not adapt to the negative bias for higher-dimensional CMIs \cite{Runge2012b,Runge2014d}. The algorithm yields different numbers of predictors for different chosen fixed thresholds or significance levels and the value should be low enough to include weak but possibly synergetic causal predictors, but high enough to limit computational complexity in the optimization step (Sect.~\ref{sec:computational_complexity}). 

\subsection{Optimal selection criteria} \label{sec:selection_criteria}
\begin{figure}[!t]
\begin{center}
\includegraphics[width=1.\columnwidth]{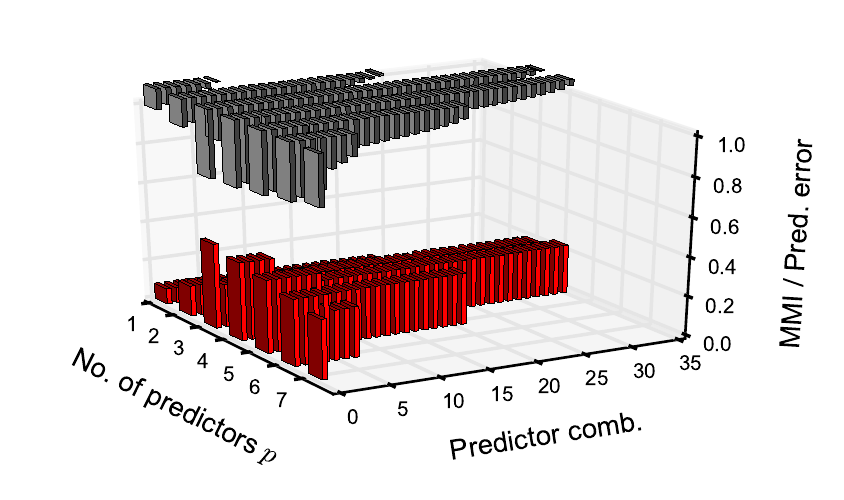}
\end{center}
\caption[]{(Color online)
Multivariate mutual information (MMI, bottom red bars) and standardized root mean squared prediction error (value proportional to lower end of grey bars at top) for all subsets of causal predictors for model~(\ref{eq:model}), for details see Sect.~\ref{sec:model_example}. For each number of predictors $p$, the number of possible combinations varies according to the binomial coefficient ${7}\choose{p}$. The predictor combinations are sorted by their prediction error. The maximum of MMI and also smaller values match the minimum prediction error very well. Note that MMI is estimated on the learning set while the prediction error is evaluated out-of-sample on the test set indicating that the bias of higher-dimensional MMIs here serves as a good proxy for overfitting as discussed in the text.
}
\label{fig:prediction_optimization_step}
\end{figure}
Once the causal predictors are determined, the optimal subset needs to be chosen (possibly containing all causal predictors). Here we also discuss a scheme using forward selection on the causal drivers (causal CMI-scheme).

For the third prediction scheme, causal CMI-selection, the forward selection ranking discussed in Sect.~\ref{sec:common_prediction_scheme} is applied not to the entire set $\mathbf{X}^-_{t+1}$, but only to the pre-selected causal predictors $\mathcal{P}_{Y_{t+h}}$. Also here, the same heuristic criterion as for the non-causal forward selection or cross-validation can be used. 

For the optimal scheme (Fig.~\ref{fig:prediction_scheme}(ii)), the MMI $\widehat{I}(\mathcal{P}^{(p)}_{t+h};Y_{t+h})$ for all subsets of the causal predictors from $\mathcal{P}_{Y_{t+h}}$ is estimated. In Fig.~\ref{fig:prediction_optimization_step} the MMI values (ensemble average) are plotted for the model example discussed later in Sect.~\ref{sec:model_example} with $|\mathcal{P}_{Y_{t+h}}|=7$ causal predictors. Even though all seven predictors are causal drivers, the \emph{estimated MMI} is highest for just three of them and even decreases for more predictors (according to Eq.~(\ref{eq:mmi_decomp}), the theoretical MMI should be maximal for seven predictors). The reason is that the estimated MMI is negatively biased for higher dimensions \cite{Kraskov2004a,Runge2012b,Runge2014d} and if the additional information contained in the predictors does not outweigh this bias, the MMI decreases. The bias of the MMI estimator, therefore, implies a penalty that avoids overfitting. In our heuristic criterion we exploit this property and simply select the subset $\mathcal{P}^{(p)}_{t+h}\subseteq \mathcal{P}_{t+h}$ with maximal estimated $\widehat{I}(\mathcal{P}^{(p)}_{t+h};Y_{t+h})$. This model-free data-based criterion could be seen as an analogue to model-based criteria like the Akaike Information Criterion (AIC) \cite{Akaike1974} where the penalty is derived from some measure of model complexity, but our criterion needs to be further investigated. We choose as the nearest-neighbor parameter in the MMI estimator $k^{\rm MMI}=k$, where $k$ is the nearest-neighbor parameter in the prediction (see next Sect.~\ref{sec:nnp}).

In our numerical experiments in Sect.~\ref{sec:numerical_experiments}, we additionally test a combination of the MMI-criterion with cross-validation: For each number of predictors $p$ we check the MMI-criterion to select the optimal combination and then use cross-validation to pick the optimal $p$. 
This approach gives always a slightly better prediction performance than using the maximum criterion alone -- at the cost of much longer computation time. 
Asymptotically, for model-based predictions the AIC criterion is equivalent to cross-validation and in our numerical experiments (see appendix Fig.~\ref{fig:robustness_length}) we also find that our heuristic MMI-criterion well matches the cross-validation choice for longer time series.
Note that our use of cross-validation treats $p$ as a tuning parameter as is typically done \cite{Hastie2009}. One could also treat the choice of a subset $\mathcal{P}^{(p)}_{t+h}$ as a tuning parameter and run steps (i)--(iii) of the prediction scheme for every fold. $\mathcal{P}^{(p)}_{t+h}$ is, however, not a numeric `tuning parameter' and different folds in the cross-validation might not contain the same subsets. As a result the variance across the folds is considerable and it is hard to find the subset with minimal cross-validation error.

\subsection{Nearest-neighbor prediction} \label{sec:nnp}
Once the optimal predictors are selected, the actual prediction is conducted here using a scheme with a fixed number of nearest neighbors $k$ \cite{Sidorowich1987}: For the optimal set of predictors $\mathcal{P}_{t+h}^{(p)}$, we first determine the distances
\begin{align}
d_{t,s} = \| \mathcal{P}^{(p)}_{t+h} - \mathcal{P}^{(p)}_{s} \|~~ \text{for all $s \in \mathcal{T}$ with $s>\tau_{\max}+h$},
\end{align}
where $\| \cdot \|$ denotes some norm. Here we apply the maximum norm as in the nearest-neighbor estimator of the (conditional) mutual information \cite{Frenzel2007}. $\tau_{\max}$ is the maximum lag used to estimate $\mathcal{P}_{t+h}^{(p)}$.
Next, we sort the distances in increasing order $d_{t,s_1}<d_{t,s_2}<\cdots$ yielding an index sequence $s_1,s_2,\ldots$. Now there are two approaches to use these distances: (i)~A fixed distance $\varepsilon$ is chosen and all points $s$ with distance smaller than $\varepsilon$ are taken into account to predict $Y_{t+h}$. Then the coarse-graining level is consistent for all sample points, but sometimes there might not be any point within a distance $\varepsilon$ \citep{Groth2001,Pompe2001}. (ii)~Here we use a fixed number of nearest neighbors which has the advantage that the same number of points contributes to a prediction making the estimate more reliable. For a chosen fixed number of nearest neighbors $k$ the future value $Y_{t+h}$ is then estimated by the conditional expectation and its prediction interval by its standard deviation:
\begin{align} \label{eq:nearest_neighbor_prediction}
\widehat{Y}_{t+h} = \frac{1}{k} \sum_{j=1}^{k} Y_{s_j},~~~~~ \widehat{\sigma}(\widehat{Y}_{t+h}) = \sqrt{\frac{1}{k} \sum_{j=1}^{k} (Y_{s_j}-\widehat{Y}_{t+h})^2}.
\end{align}
Another option, instead of the expectation, is to fit an autoregressive model giving a \emph{local-linear prediction} \citep{Sidorowich1987}. The summation can also be weighted with a function of the distance of the neighbors, different norms, or kernel-based methods can be chosen \cite{Hastie2009}.

For the number of nearest neighbors $k$, we use $k=k^{\rm CMI/MMI}$ where $k^{\rm CMI/MMI}$ is the nearest-neighbor parameter in the estimation of CMI or MMI in the selection schemes. This choice approximately yields a consistent level of coarse-graining in the information-theoretic selection step and the nearest-neighbor prediction. Alternatively, at the cost of computation power, one can also utilize cross-validation for this choice \cite{Hastie2009}. 
The number of nearest neighbors needs to be balanced to guarantee that only nearby values are used as predictors, but still enough values are available to confidently estimate $Y_{t+h}$ and possibly the prediction interval. The value will typically strongly depend on the data.
As a skill metric we compute the standardized root mean squared error
\begin{align} \label{eq:prediction_skill}
{\rm SRMSE} &= \sqrt{\frac{\frac{1}{|\mathcal{T}_{\rm test}|} \sum_{t\in \mathcal{T}_{\rm test}} (Y_{t+h}-\widehat{Y}_{t+h})^2}{\sigma^2_Y}  },
\end{align}
where $\sigma^2_Y$ is the variance of $Y$ in the testing set $\mathcal{T}_{\rm test}$.

\section{Computational complexity} \label{sec:computational_complexity}
\begin{figure}[!t]
\begin{center}
\includegraphics[width=1.\columnwidth]{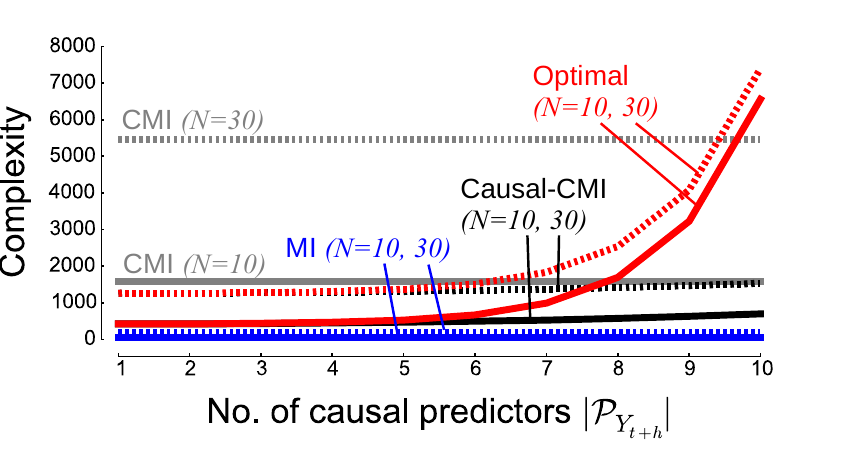}
\end{center}
\caption[]{(Color online)
Scaling of computational complexity with the cardinality of the estimated causal predictors $|\mathcal{P}_{Y_{t+h}}|$ for $\tau_{\max}=2$. Solid lines will be used for $N=10$, whereas dashed lines denote $N=30$. 
The curve for the causal CMI-selection scheme (black) is only slightly different from the constant curve for the MI-selection scheme (blue). 
The causal schemes have the advantage that their complexity rather depends on how many causal drivers are found, while the complexity of the MI- and CMI-selection schemes depends on the number of processes and the maximum lag. For the non-causal CMI-selection scheme (grey lines) we fixed the maximum number of predictors to $p_{\max}=10$. To include the complexity due to the preselection step in the causal schemes we added $420$ ($1260$) for $N = 10$ ($N = 30$) for these schemes according to Eq.~(\ref{eq:algo_complexity}) for $n_0 = 2$ and $n_i = 3$ as in our numerical experiments (see Sect.~\ref{sec:numerical_experiments}).
The plot shows that if the number of causal predictors can be reduced below $8$ ($10$) here, the optimal scheme even takes less computation time than the non-causal CMI-selection scheme.
}
\label{fig:computational_complexity}
\end{figure}

An important criterion for the practical applicability of the different prediction schemes discussed in Sects.~\ref{sec:common_prediction_scheme} and \ref{sec:optimal_prediction_scheme} is their respective computational complexity. The estimator for the (C)MI $I(X;Y|Z)$ employed here (nearest-neighbor technique with maximum norm \cite{Frenzel2007}) has a computational complexity of $\mathcal{O}(T^2 (D_X+D_Y+D_Z))$ \cite{Kraskov2004a}, where $T$ is the time series length and $D$ the dimensionality of the respective variable. Fast neighbor searching algorithms can further reduce the dependency on $T$, but here we are interested in the relative complexity of the different predictor selection schemes and, therefore, only consider the linear scaling with the number of dimensions. In this case the first prediction scheme, MI-selection, clearly is the cheapest option. For a $N$-dimensional process $\mathbf{X}$, this procedure involves just $N(\tau_{\max}+1)$ estimates of MIs with a dimensionality of $D_X=1,\,D_Y=1,\,D_Z=0$. 

The second scheme, CMI-forward selection, is more demanding the more possible predictors are included. For cross-validation, a maximum number of $p_{\max}$ predictors has to be selected, increasing the dimension to maximally $D_X=1,\,D_Y=1,\,D_Z=p_{\max}-1$ due to more conditions in Eq.~(\ref{eq:cmi_decomp}).
The CMI-forward selection then involves 
\begin{align*}
&\sum_{p=0}^{p_{\max}-1} (N(\tau_{\max}+1)-p) \\
&= \frac{1}{2} p_{\max}(1-p_{\max}+2 N (1+\tau_{\max} ))
\end{align*}
estimates of CMI with iteratively increasing dimensionality. Then the complexity of the CMI-selection scheme scales as
\begin{align*}
&\sum_{p=0}^{p_{\max}-1} (N(\tau_{\max}+1)-p) (p+2) \\
&=\frac{1}{6} p_{\max} \left(5-3 p_{\max}-2 p_{\max}^2+3 N (3+p_{\max}) (1+\tau_{\max} )\right)
\end{align*}
for $p_{\max}<N(\tau_{\max}+1)+1$, i.e., with a high linear dependency on $N(\tau_{\max}+1)$ and a polynomial dependency $\sim p_{\max}^3$.
Note that for a $m$-fold cross-validation step using nearest-neighbor prediction an additional computational complexity of 
\begin{align*}
&m \cdot T \frac{T}{m}\sum_{p=1}^{p_{\max}} (1+p) =\frac{1}{2} T^2 p_{\max} (3+p_{\max})
\end{align*}
has to be added.

In Fig.~\ref{fig:computational_complexity} we compare the complexity of the different prediction schemes for $N=10$ and $\tau_{\max}=2$ as in our numerical experiments in Sect.~\ref{sec:numerical_experiments}. While one can fix $p_{\max}$ to a small number for which nearest-neighbor predictions yield acceptable results, the main problem here is that the computation time quickly increases with $N(\tau_{\max}+1)$ (linear, but with a large pre-factor). One advantage of a causal pre-selection step is to reduce this number before the computationally expensive selection procedure is invoked.

Since typically the set of causal predictors is small, i.e., $|\mathcal{P}_{Y_{t+h}}| \ll N(\tau_{\max}+1)$, the third prediction scheme, causal CMI-selection, has a drastically smaller computational complexity than the non-causal CMI-selection scheme if the additional complexity due to the causal algorithm is not that large. If we rank all causal predictors (not limiting to some $p_{\max}$ as in the non-causal scheme), the complexity scales as $\frac{1}{6} |\mathcal{P}_{Y_{t+h}}| (|\mathcal{P}_{Y_{t+h}}| + 1) (|\mathcal{P}_{Y_{t+h}}|+5)$ where $|\mathcal{P}_{Y_{t+h}}|$ denotes the number of causal predictors. In Fig.~\ref{fig:computational_complexity}, the complexity (black lines) shows only a very moderate increase with $|\mathcal{P}_{Y_{t+h}}|$.

For the optimal scheme the computational complexity grows exponentially as $2^{|\mathcal{P}_{Y_{t+h}}|-1} (2+|\mathcal{P}_{Y_{t+h}}|)-1$. However, Fig.~\ref{fig:computational_complexity} shows that if the number of causal predictors can be restricted, the optimal scheme even takes less computation time than the non-causal CMI-selection scheme.
The number of causal predictors can be reduced by adjusting the significance level $\alpha$ or fixed threshold $I^*$ in the conditional independence tests of the causal pre-selection algorithm. Most important, the causal scheme's complexity only scales with the number of causal drivers and \emph{not} directly with the number of processes $N$ or the maximal lag $\tau_{\max}$ as the non-causal schemes (dashed lines in Fig.~\ref{fig:computational_complexity}). The dependence of the causal schemes on $N$ and the maximal lag $\tau_{\max}$ is only via the algorithm.

The additional time complexity of the causal algorithm varies with the graph structure. The number of iterations can be limited by starting with a higher number of initial conditions $n_0$ and limiting the maximum dimensionality $n_{\max}$. This number determines up to which dimension of $\mathcal{P}^{n,i}_{Y_{t+h}}$ the conditional independence is checked. Also the number of combinations $n_i$ in the $i$-loop can be restricted. In a worst case scenario where the spurious links only vanish if the maximum number of conditions is used, the computational complexity scales as
\begin{align*}
&N(\tau_{\max}+1) \left(2 + \sum_{n=n_0}^{n_{\max}} \sum_{i=0}^{n_{i}-1} (2+n) \right) \nonumber\\
&= N(\tau_{\max}+1) \left(2 + \frac{n_i}{2} (1+n_{\max}-n_0)(4+n_{\max}+n_0) \right).
\end{align*}
However, for sparse graphs and the conditional set being efficiently chosen \cite{Runge2012prl}, typically links get removed already for an $n_0$-dimensional condition with a complexity of
\begin{align} \label{eq:algo_complexity}
N(\tau_{\max}+1) (2 + n_i (2+n_0))\,,
\end{align}
and $n_i=1$ or $2$. This is also confirmed in numerical experiments in \cite{Runge2012prl} and Sect.~\ref{sec:numerical_experiments}. Often the complexity is even lower because the MI value in the first iteration is already non-significant.

\section{Model example} \label{sec:model_example}
\begin{figure}[!t]
\begin{center}
\includegraphics[width=1.\columnwidth]{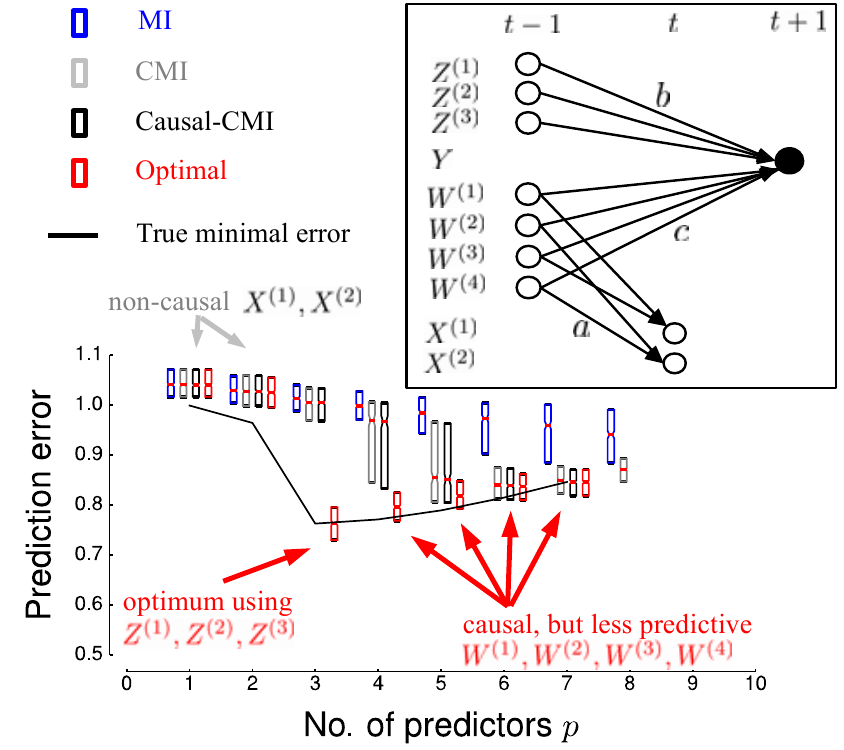}
\end{center}
\caption[]{(Color online) Comparison of prediction schemes for an ensemble of $500$ realizations of model~(\ref{eq:model}) for $a=0.4,\,b=2,\,c=0.4,\,\sigma=0.5$ with time series graph given in the inset (learning set length $500$, test set length $125$, nearest-neighbor prediction with $k=10$ neighbors, (C)MIs estimated from the learning set with parameter $k^{\rm algo}=50$ in the algorithm and $k^{\rm CMI/MMI}=10$ for the optimization). The box and whiskers plots give the ensemble median and the interquartile range of the standardized root mean squared prediction errors in the test set for each iteration step $p$ in the four schemes (from left to right: MI, CMI, causal-CMI, optimal scheme). The black line gives the median of the true minimal prediction error obtained by minimizing the out-of-sample prediction error for each number of predictors $p$ taken from the true causal drivers. For $p=3$, only the optimal scheme (red) selects the best (synergetic) predictors $Z_{t-1}^{(1)},Z_{t-1}^{(2)},Z_{t-1}^{(3)}$ and reaches the minimum possible error while the causal forward selection (black) first picks one of the less predictive $W_{t-1}^{(\cdot)}$ and the pure forward selection (grey) and MI-based schemes first pick the two non-causal drivers $X_{t}^{(1)},X_{t}^{(2)}$. The non-optimal schemes include the synergetic predictors only when the higher dimensionality is already worsening the prediction due to overfitting.}
\label{fig:model_example}
\end{figure}
The following nonlinear discrete-time stochastic delay equation provides an illustrative example where a simple MI- or CMI-forward selection procedure yields non-causal variables that deteriorate a prediction. Consider
\begin{align} \label{eq:model}
Y_{t+1} &= c \sum_{i=1}^{4} W^{(i)}_{t-1} + b\, \prod_{i=1}^{3}Z^{(i)}_{t-1} + \sigma \eta^Y_{t+1} \nonumber \\
X^{(1)}_t &= a \left(W^{(1)}_{t-1}+W^{(3)}_{t-1} \right) + \eta^{X^{(1)}}_t \nonumber \\
X^{(2)}_t &= a \left(W^{(2)}_{t-1}+W^{(4)}_{t-1} \right) + \eta^{X^{(2)}}_t 
\end{align}
where the causal drivers $W^{(\cdot)}$, $Z^{(\cdot)}$, and $\eta^{\cdot}$ are independent Gaussian processes with zero mean and unit variance [$\sim\mathcal{N}(0,1)$]. This model illustrates why the MI and CMI prediction schemes fail to yield good predictions due to (1)~selecting non-causal predictors and (2)~missing out synergetic predictors. Here the $Z^{(\cdot)}_{t-1}$ are synergetic causal drivers, which -- for certain parameters $b,\,c$ -- are \emph{individually} less predictive than the drivers $W^{(\cdot)}_{t-1}$. But selected all \emph{together}, their combined information $I\left( (Z^{(1)}_{t-1},Z^{(2)}_{t-1},Z^{(3)}_{t-1})\,;\,Y_{t+1}\right)$ is much larger than the single mutual informations. This synergetic effect can only be detected if all subsets of causal drivers are tested in a globally optimal scheme. In the following we analyze why the MI- and CMI-selection schemes fail to provide good predictions for such cases.

Regarding the problem of selecting non-causal drivers, for certain parameter combinations of $(a,\,b,\,c)$ the mutual information $I(X^{(i)}_{t}\,;\,Y_{t+1})$ is larger than any $I(W^{(\cdot)}_{t-1}\,;\,Y_{t+1})$ or $I(Z^{(\cdot)}_{t-1}\,;\,Y_{t+1})$ for all $i$. 
The non-causal schemes based on iteratively selecting predictors with maximal MI or CMI (blue and grey box plots in Fig.~\ref{fig:model_example}) will, therefore, choose a non-causal $X^{(\cdot)}_{t}$ prior to the true causal predictors $W^{(\cdot)}_{t-1}$ and $Z^{(\cdot)}_{t-1}$. Since the predictors $W^{(\cdot)}_{t-1}$ have the largest MI after the $X^{(\cdot)}_{t}$, these are included next in the MI-selection scheme.
In the CMI-forward selection scheme, on the other hand, the synergetic variables $Z^{(\cdot)}_{t-1}$ are selected after the second iteration step. This leads to a drastic decrease in the prediction error at $p=5$ (grey box plot in Fig.~\ref{fig:model_example}). 
The problem is that now the dimension of the predictors is already $5$ and the two spuriously causal variables $X^{(\cdot)}_{t}$ deteriorate the prediction.

The causal pre-selection avoids this pitfall. The black and red box plots in Fig.~\ref{fig:model_example} denote the schemes based on causal predictors using forward selection (black) and the optimal scheme (red). Also for causal drivers the forward selection scheme is not optimal because the selection of one predictor at a time fails for synergetic cases and selects the weak drivers $W^{(\cdot)}$ prior to the synergetic drivers $Z^{(\cdot)}$. Only the optimal scheme correctly identifies these drivers for the dimension $p=3$ and reaches the minimal prediction error possible for this model (black line) for each $p$. In Fig.~\ref{fig:prediction_optimization_step} we show that the three synergetic drivers $Z^{(\cdot)}$ are chosen with our heuristic optimal criterion since they have the largest MMI with the target variable.

\section{Numerical experiments} \label{sec:numerical_experiments}
\begin{figure}[!t]
\begin{center}
\includegraphics[width=1.\columnwidth]{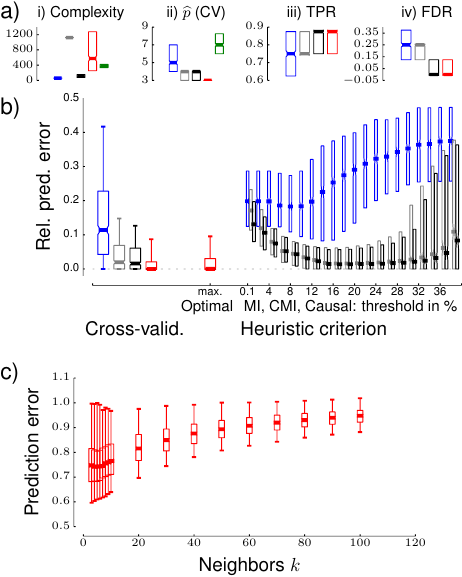}
\end{center}
\caption[]{(Color online)
Results of numerical experiments for an ensemble of $500$ trials of the synergetic model class~(\ref{eq:model_numerical_experiments}) with  time series length $T=500$ ($125$ in the test set) for the four different prediction schemes (from left to right in the panels: MI in blue, CMI in gray, causal-CMI in black, optimal in red). 
(a) The four box plots show the ensemble interquartile range of (i)~computational complexity (the green box on the right shows the added complexity due to the causal algorithm), (ii)~the range of numbers of predictors $\widehat{p}$ selected by cross-validation (CV, here the green bar shows the number of causal predictors $|\mathcal{P}_{Y_{t+h}}|$ in the pre-selection step), (iii)~the true positive rate (TPR) and (iv)~false discovery rate (FDR). The latter are evaluated for each scheme at $p=8$, corresponding to the true number of causal drivers (or less if fewer causal drivers are estimated in the pre-selection step). 
(b)  Box plots showing the median and interquartile range of the prediction error \emph{relative} to the true minimal error obtained by minimizing the out-of-sample prediction error over all subsets of true causal drivers. On the left are the results if cross-validation is used to optimize $p$ for each scheme (whiskers show the 5\% and 95\% quantiles). The red box in the center shows the result for the heuristic optimality criterion. The range of boxes on the right shows the results for different thresholds $\lambda$ for the other schemes (only interquartile range). 
(c) Box and whiskers plots (showing the 5\% and 95\% quantiles) for the absolute prediction error of the optimal scheme at the cross-validated choice of $p$ for different numbers of nearest neighbors $k$.
}
\label{fig:numerical_experiments}
\end{figure}
\begin{figure}[!t]
\begin{center}
\includegraphics[width=1.\columnwidth]{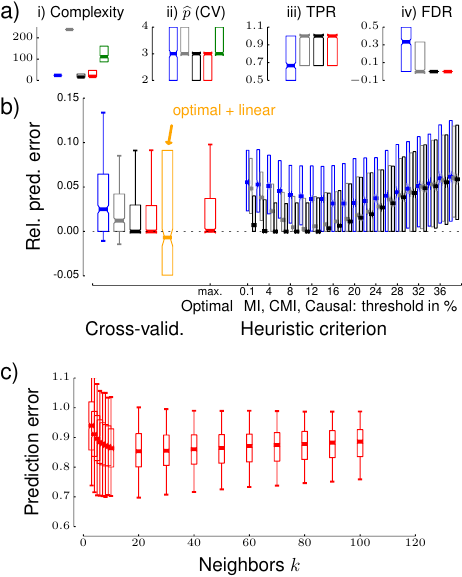}
\end{center}
\caption[]{(Color online) Numerical experiments on a class of non-synergetic, but still nonlinearly coupled generalized additive models as analyzed in the supplement of Ref.~\cite{Runge2012prl}. Parameters as in Fig.~\ref{fig:numerical_experiments}, but with $N=4$ processes and $p_{\max}=6$. The orange box plot in (b) shows the relative prediction error if the optimal predictors from the model-free selection scheme are used in conjunction with the linear auto-regressive prediction model~(\ref{eq:linear_prediction}) (only the interquartile range shown).}
\label{fig:prl_model}
\end{figure}
We next compare the four schemes including the causal pre-selection algorithm on a larger class of synergetic nonlinear discrete-time stochastic processes with different coupling configurations:
\begin{align} \label{eq:model_numerical_experiments}
\mathbf{X}^{(1)}_{t+1} &= c\, g_{\rm lin}^{(1)} (\mathbf{X}^{-}_{t+1}) + b\,g_{\rm syn} (\mathbf{X}^{-}_{t+1}) + \eta^{(1)}_{t+1} \nonumber \\
\mathbf{X}^{(j)}_{t+1} &= a\, g_{\rm lin}^{(j)} (\mathbf{X}^{-}_{t+1})+  \eta^{(j)}_{t+1} ~~~~\text{for $j=2,\ldots,\,N$}\,,
\end{align}
where we are interested in predicting $\mathbf{X}^{(1)}_{t+1}$ (i.e., $h=1$). The linear function $g_{\rm lin}^{(1)}$ is simply the sum of $5$ randomly chosen subprocesses $\mathbf{X}^{(\cdot)}_{t-\tau}$ (excluding process $\mathbf{X}^{(1)}$) at random lags $0\leq \tau \leq 2$. The nonlinear function $g_{\rm syn}$, on the other hand, is the product of $3$ randomly chosen subprocesses (excluding process $\mathbf{X}^{(1)}$ and the ones already included in the linear term). The other subprocesses for $j>1$ are linearly driven by $2$ other randomly chosen subprocesses, also at random lags. The coefficients are fixed to $a=0.4,\,b=2,\,c=0.4$, and $N=10$. With this setup we generate an ensemble of $500$ realizations.

We run the four schemes at different choices of the heuristic parameter $\lambda$ and using cross-validation checking up to $p_{\max}=8$ predictors in the non-causal schemes. For the causal schemes, we use cross-validation for all estimated causal predictors (up to maximally $p_{\max}=8$, ranked by their CMI value in the algorithm \cite{Runge2012prl}). The causal drivers are estimated using the algorithm parameters $n_0=2,n_{\max}=3,\,n_i=3$, and $\tau_{\max}=2$ with a fixed significance threshold $I^*=0.004$ (analyses for other thresholds are shown in the appendix).

In Fig.~\ref{fig:numerical_experiments}, we show various statistics comparing the computational complexities and the prediction errors. Here the number of possible predictors is $N(\tau_{\max}+1)=30$ yielding a computational complexity of $60$ for the MI-selection scheme (blue) and $1124$ for the CMI-selection scheme (grey) using $p_{\max}=8$. The causal algorithm reduces the number of possible predictors to about $7$ (median). This corresponds to a true positive rate (TPR) of roughly $0.9$ (there are $8$ true drivers in model~(\ref{eq:model_numerical_experiments}), but several are only weakly driving) and a zero false discovery rate (FDR), while the MI- and CMI-selection schemes detect fewer causal drivers and much more false positives. Fewer predictors result in a lower computational complexity for the causal prediction schemes. The causal CMI-selection scheme runs extremely fast (black) and the complexity of the optimal scheme (red) strongly varies among the different realizations, since it depends exponentially on how many causal predictors are pre-selected (Fig.~\ref{fig:computational_complexity}), but still typically even stays below the non-causal CMI-selection scheme. Using cross-validation, the MI-selection scheme uses typically (median) $p=5$, the CMI-selection schemes both $p=4$, and the optimal scheme only $p=3$ out of the $8$ true causal drivers for this model. 

Finally, the relative prediction errors show that only the optimal scheme reaches the lowest possible errors with a median of zero relative error and even 90\% of the ensemble below an error of $0.1$. This demonstrates the large improvements due to the global optimization scheme that is only possible after reducing the set of variables to the few causal predictors. 

The aforementioned results have been obtained using cross-validation to select the optimal $p$. The computationally cheaper alternative using a heuristic criterion here yields drastic differences in the prediction performance depending on the choice of $\lambda$. While here values in the range $\lambda=20\%\ldots 30\%$ give good results, in another experiment (Fig.~\ref{fig:prl_model}) we found good predictions only for $\lambda=10\%\ldots 15\%$ making it hard to provide rules of thumb in practical applications. In the appendix we show that also the length of the time series results in different optimal ranges for $\lambda$. On the other hand, for the optimal scheme the heuristic choice leads to almost the same minimal errors as in cross-validation.  

To test the robustness of our results, we also compare the prediction schemes on a class of non-synergetic, but still nonlinearly coupled models (generalized additive models \cite{Hastie2009} with $N=4$ processes and polynomials of linear and quadratic degree) as analyzed in the supplement of Ref.~\cite{Runge2012prl}. For each ensemble member, we choose as a target variable the one with the largest sum of `incoming' coefficients (absolute values). The results shown in Fig.~\ref{fig:prl_model} demonstrate that for this case also the causal CMI-selection scheme reaches optimal prediction errors. In the appendix, we show that the optimality is robust also for different significance thresholds and other time series lengths.

In Figs.~\ref{fig:numerical_experiments}(c) and Fig.~\ref{fig:prl_model}(c) we evaluate the prediction for different phase space resolutions. To this end we use the causal predictors and run steps (ii) and (iii) of the prediction scheme from Fig.~\ref{fig:prediction_scheme} (using cross-validation to choose $\widehat{p}$) for varying nearest-neighbor parameters $k$ and MMI estimation parameter $k^{\rm MMI}=k$, both set to the same value for consistency. While in the first ensemble (Fig.~\ref{fig:numerical_experiments}(c)) the error is minimal for very few neighbors and sharply rises if too many neighbors are used, in the second ensemble (Fig.~\ref{fig:prl_model}(c)) too few neighbors yield worse results. In practice, the choice will very much depend on the process under study, but here we use a value in the range $k=5\ldots 10$ which constitutes a balance between local information and enough neighbors to reliably estimate $\widehat{Y}_{t+h}$.

\section{Model-free selection combined with model-based prediction} \label{sec:optimal_plus_model}
Up to now we have stayed in a model-free framework with information-theoretic optimal selection of predictors and a nearest-neighbor prediction. While nearest-neighbor prediction is a flexible method that will adapt to any function $f$ in Eq.~(\ref{eq:evolution}), it will in many cases be outperformed by a model-based prediction -- if the right model class is chosen. If a misspecified model is chosen for variable selection and fitting, it might miss out nonlinear combinations of predictors. For example, in our synergetic model~(\ref{eq:model}) a linear selection method would only include the weakly predictive variables $W^{(\cdot)}$ and largely miss out the highly predictive variables $Z^{(\cdot)}$. The functional dependency on the $W^{(\cdot)}$ is, on the other hand, much better fitted with a linear model than with nearest neighbors.

To take advantage of improved model-based predictions and at the same time not miss out synergetic predictor combinations, we propose to apply our optimal predictor selection scheme and conduct the final prediction step by fitting a model on the optimal set of predictors. Here we demonstrate this approach on the non-synergetic model ensemble from Ref.~\cite{Runge2012prl}. To predict $Y_{t+h}$ from the optimal subset of predictors $\mathcal{P}^{(\widehat{p})}_{Y_{t+h}}$ (chosen by cross-validation or the heuristic criterion), we use the ordinary least squares regression technique. Then the prediction interval is given by the variance $\widehat{\sigma}^2_{\varepsilon}$  of the regression residual plus the errors in the estimated regression coefficients $\widehat{\mathbf{B}}$ \cite{Brockwell2002}:
\begin{align} \label{eq:linear_prediction}
\widehat{Y}_{t+h} = \mathcal{P}^{(\widehat{p})}_{t+h} \widehat{\mathbf{B}},~~~~~ \widehat{\sigma}(\widehat{Y}_{t+h}) = \sqrt{\widehat{\sigma}^2_{\varepsilon} + \sum_{i=1}^{\widehat{p}} \widehat{\sigma}(\widehat{\mathbf{B}_i})^2 (X^{(i)})^2  }.
\end{align}
In Fig.~\ref{fig:prl_model}(b) the results for this approach are shown as the orange box plot. The prediction improvement varies strongly for the different realizations which include nonlinear and linear drivers. About half of the realizations are better fitted using the optimized linear approach with prediction improvements of up to 10\% compared to the nearest-neighbors prediction. More advanced techniques such as generalized additive models can further improve a prediction \cite{Hastie2009}. In addition to facilitating the prediction task, the knowledge of the functional forms of dependencies can also help to better understand coupling mechanisms. 

\section{Predicting ENSO} \label{sec:enso_prediction}
\begin{figure}[!t]
\begin{center}
\includegraphics[width=1.\columnwidth]{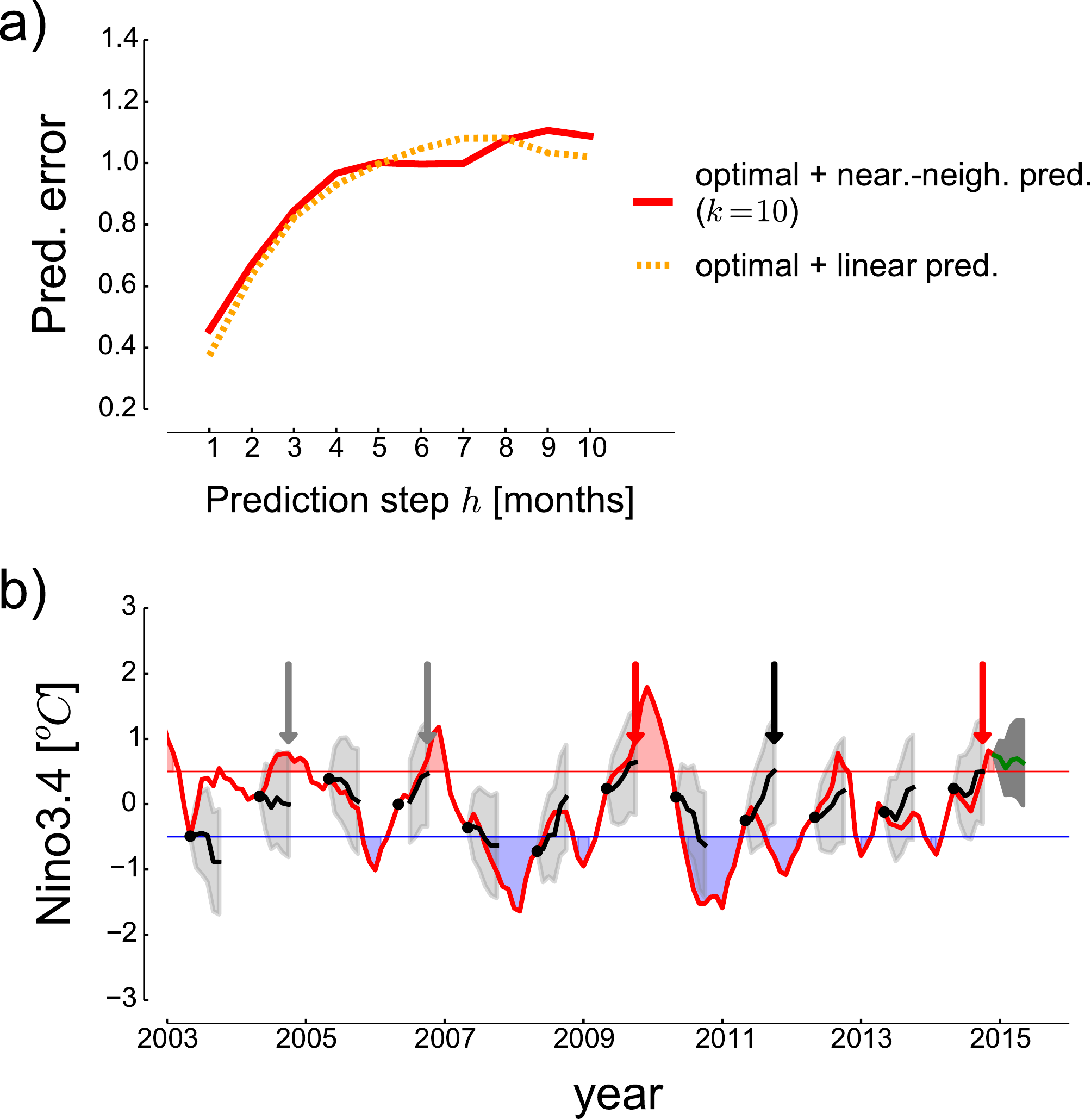}
\end{center}
\caption[]{(Color online) Prediction of the El Ni\~no Southern Oscillation (ENSO) index Nino3.4 in the period 2003-2014 (up to December) using 1951-2002 as a learning set, causal algorithm run with significance threshold $I^*=0.03$ testing up to $\tau_{\max}=12$ months. (a)~Prediction error using nearest-neighbor prediction (solid red line) and linear prediction (dotted orange line) versus prediction step $h$. For both approaches the same optimal predictors obtained from the model-free scheme with cross-validated (5-fold within the learning set) choice of predictors are used. (b)~Nino3.4 index with El Ni\~no and La Ni\~na events marked in red and blue, respectively. The black lines denote selected hindcasts  and their $1\sigma$-prediction intervals (grey) using the linear prediction. The dots mark the starting time $t$ in May of each year, and the predicted values range from June ($h=1$) to October ($h=5$). The arrows mark correct (red), missed (grey) and false (black)  hindcasts of the Nino3.4 index exceeding  \SI{0.5}{\degreeCelsius} with more than 49\% probability. The green line marks a real forecast starting in December 2014 giving a probability for the Nino3.4 index to stay above \SI{0.5}{\degreeCelsius} of 55-70\% for the months until May 2015.}
\label{fig:enso_prediction}
\end{figure}

The combined framework developed in the last section is now illustrated on a sea-surface temperature index of the El Ni\~no Southern Oscillation (ENSO) in the tropical Pacific which has been the focus of prediction research for many decades due to its far-reaching climatic and economic impacts \cite{Cane1986,Barnston2012}. The Nino3.4 index is defined as the average sea-surface temperature over the region \SI{5}{\degree}N-\SI{5}{\degree}S,  \SI{170}{\degree}-\SI{120}{\degree}W \cite{Al2008}. As another possible predictor variable, we use an atmospheric index based on sea-level pressure, the Southern Oscillation Index (SOI), which is computed from the surface air pressure difference between Tahiti and Darwin, Australia.

In Fig.~\ref{fig:enso_prediction}(a), we employ the model-free causal algorithm and predictor selection (steps (i)-(ii) in Fig.~\ref{fig:prediction_scheme}, here optimized using cross-validation) to obtain the optimal predictors and compare the skill of the nearest-neighbor and the linear prediction using the auto-regressive model~(\ref{eq:linear_prediction}) fitted on the optimal predictors. Trained on the period 1951-2002, we test the prediction on the last decade 2003-2014. From the 24 possible predictors, the optimal predictor for $h=1$ month is only Nino3.4 at one month lag, while for $h=2$ months the three predictors $({\rm Nino3.4}_{t},{\rm SOI}_{t},{\rm SOI}_{t-9})$ are relevant indicating that the atmospheric coupling, including a long memory, constitutes an important predictive mechanism. Here, the linear auto-regressive model significantly reduces the prediction error by about $0.05-0.1$ compared to the nearest-neighbor approach using the same predictors, at least for a few months ahead. For steps larger than 5 months, the error in both approaches quickly reaches 1 which implies that the prediction is merely a persistence forecast. The better linear prediction is a sign that exploiting the nonlinearities in Nino3.4 \cite{Dommenget2013} does not improve the prediction much while the linear fit using the optimal predictors better harnesses the linear drivers of ENSO -- at least on these time scales \cite{Gamez2004a}. 

To give an impression of selected predictions from the linear model (actually hindcasts), we show in Fig.~\ref{fig:enso_prediction}(b) the predictions up to 5 months ahead starting from May in each year. The important onsets of El Ni\~no events are determined by expert assessment, but one definition is the 3-month-running-mean smoothed Nino3.4 index exceeding \SI{0.5}{\degreeCelsius}, here marked by a red line (La Ni\~nas, where the index decreases below \SI{-0.5}{\degreeCelsius} are marked in blue). With our hindcasts starting in May of each year, one can compute the probability of an El Ni\~no event as the part of the prediction distribution exceeding \SI{0.5}{\degreeCelsius} (assuming a Gaussian distribution with mean and standard deviation given by Eq.~(\ref{eq:linear_prediction})). If this probability is larger than 49\% for any of the 5 months ahead (until October), we predict an El Ni\~no event. With this scheme we would have correctly predicted the moderate El Ni\~no event in 2009 and the onset of the weak El Nino in current 2014-2015 season (red arrows), but missed the weak events of 2004 and 2006 (grey arrows, the latter being almost predicted with a probability of 48\%). On the other hand, in 2011 (black arrow) a false alarm is given. The overall weak predictability of the recent El Ni\~no events is also found in other studies using statistical as well as physical model predictions \cite{Barnston2012} and suggests that the mechanism of ENSO could be changing.
Finally, our real forecast (green line) starting from December 2014 suggests that the weak current El Ni\~no condition persists with a probability of 55-70\% for the months until May 2015.

\section{Discussion and conclusions}
In this article we have shown that the combinatorial explosion to search for globally optimal subsets of predictors can be overcome by restricting the search to causal drivers. Globally optimal predictors detect also synergetic mechanisms where the combination of multiple predictors strongly improves a prediction. Analytical considerations and numerical experiments indicate that such an approach is superior to schemes using MI-ranking or forward selection with conditional mutual information.  
Another advantage is that the computational complexity only scales with the number of causal predictors and not directly with the number of processes included in the analysis. If the set of causal predictors is not that large, the optimal scheme is even computationally less expensive than the non-causal CMI-selection scheme.
To determine the optimal size of this set, we have found that a parameter-free heuristic criterion performs almost as good as a computationally much more demanding cross-validation.

Note that, even though theoretically only causal drivers can yield optimal predictions, non-causal variables could still be better predictors. Consider the case that a very high-dimensional process $\mathbf{W}$ drives $Y$ and $X$. Then the prediction of $Y$ from the causal drivers $\mathbf{W}$ is deteriorated due to the curse of dimensionality for finite samples, while the non-causal process $X$ could potentially better aggregate this information. The same effect also explains why in Fig.~\ref{fig:model_example} the CMI prediction using the non-causal $X^{(\cdot)}_{t}$ together with the synergetic drivers has a slightly smaller prediction error than the causal CMI-selection scheme for $p=5$ (grey box plot). 

While we propose the model-free selection of predictors for processes where the underlying mechanisms are poorly understood, the actual prediction can be much improved using suitable model-based techniques compared to a pure model-free nearest-neighbor prediction. This approach
combines the advantage of a model-free approach to detect relevant variables with the smaller prediction variance of model-based methods and can also be used to better understand coupling mechanisms. The application of this combined approach significantly improves the prediction of an ENSO index compared to a nearest-neighbor scheme.
The combined approach can be further improved by optimizing the number of predictors $\widehat{p}$ with a different criterion than the model-free criteria discussed in Sect.~\ref{sec:selection_criteria}. Especially linear models can harness much more predictors before the problem of overfitting becomes severe.

Here the scope of application was the prediction of future values of a time series. In a forthcoming paper we will investigate how the scheme can be adapted if, for example, only forecasts for the emergence of extreme events like El Ni\~nos \cite{Ludescher2013} are needed.
A \emph{Python} script to estimate the causal predictors can be obtained from the author's website at \texttt{www.pik-potsdam.de/members/jakrunge}.

\section*{Acknowledgments}
We acknowledge the financial support by the German National Academic Foundation (Studienstiftung des deutschen Volkes), the Humboldt Graduate School, and the German Federal Ministry of Education and Research (Young Investigators Group CoSy-CC$^2$, grant no. 01LN1306A).

\section*{Appendix}

\renewcommand{\theequation}{A\arabic{equation}}
\setcounter{equation}{0}
\setcounter{section}{0}
\section{Robustness}
\begin{figure*}[!t]
\begin{center}
\includegraphics[width=.8\textwidth]{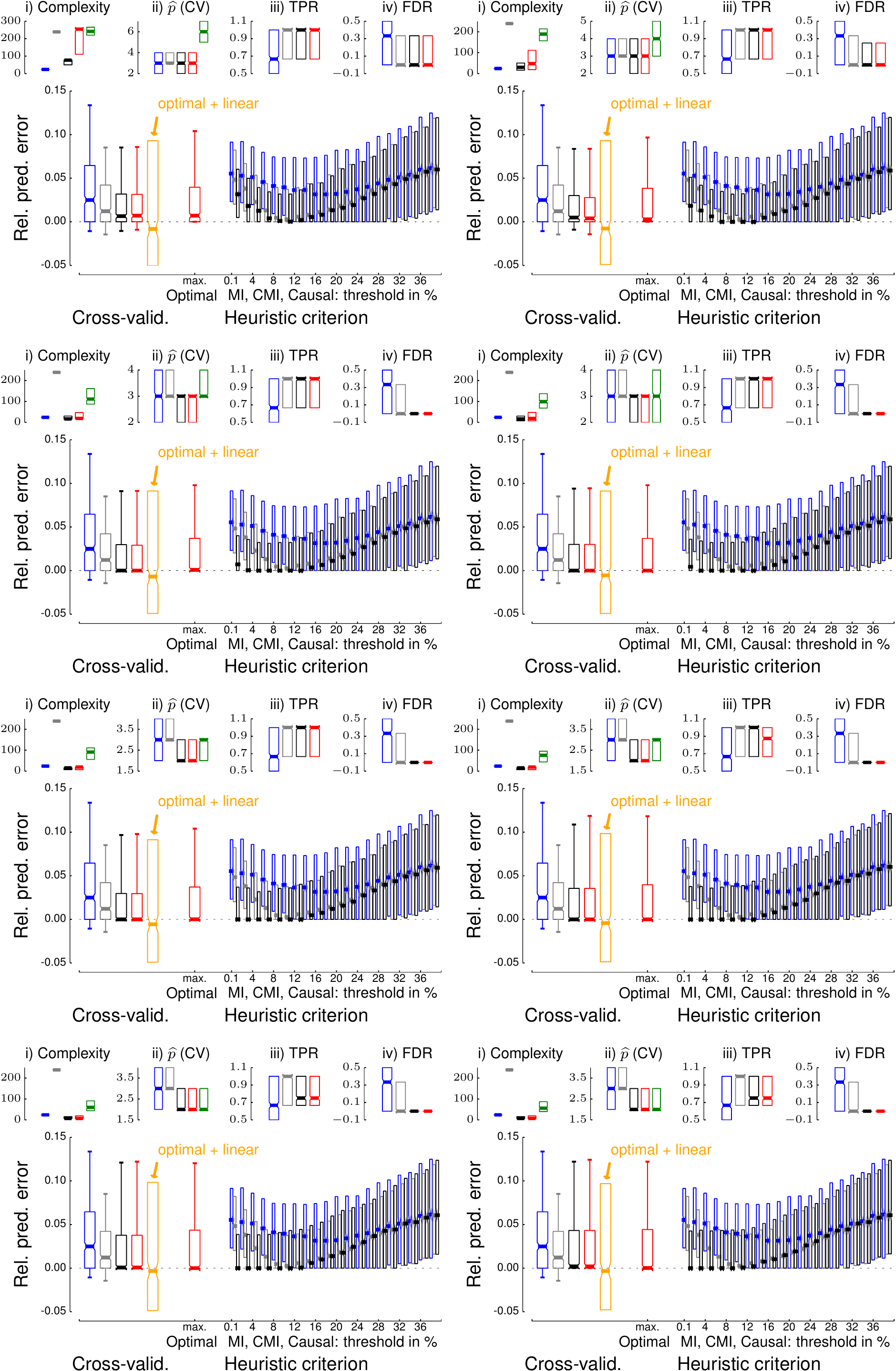}
\end{center}
\caption[]{(Color online) As in Fig.~\ref{fig:prl_model}(a,b), but for a larger range of significance thresholds $I^*=0.001,\,0.002,\,0.004,\,0.005,\,0.006,\,0.008,\,0.009,\,0.01$ (row-wise from top left to bottom right).}
\label{fig:robustness_sig}
\end{figure*}
\begin{figure*}[!t]
\begin{center}
\includegraphics[width=1.\textwidth]{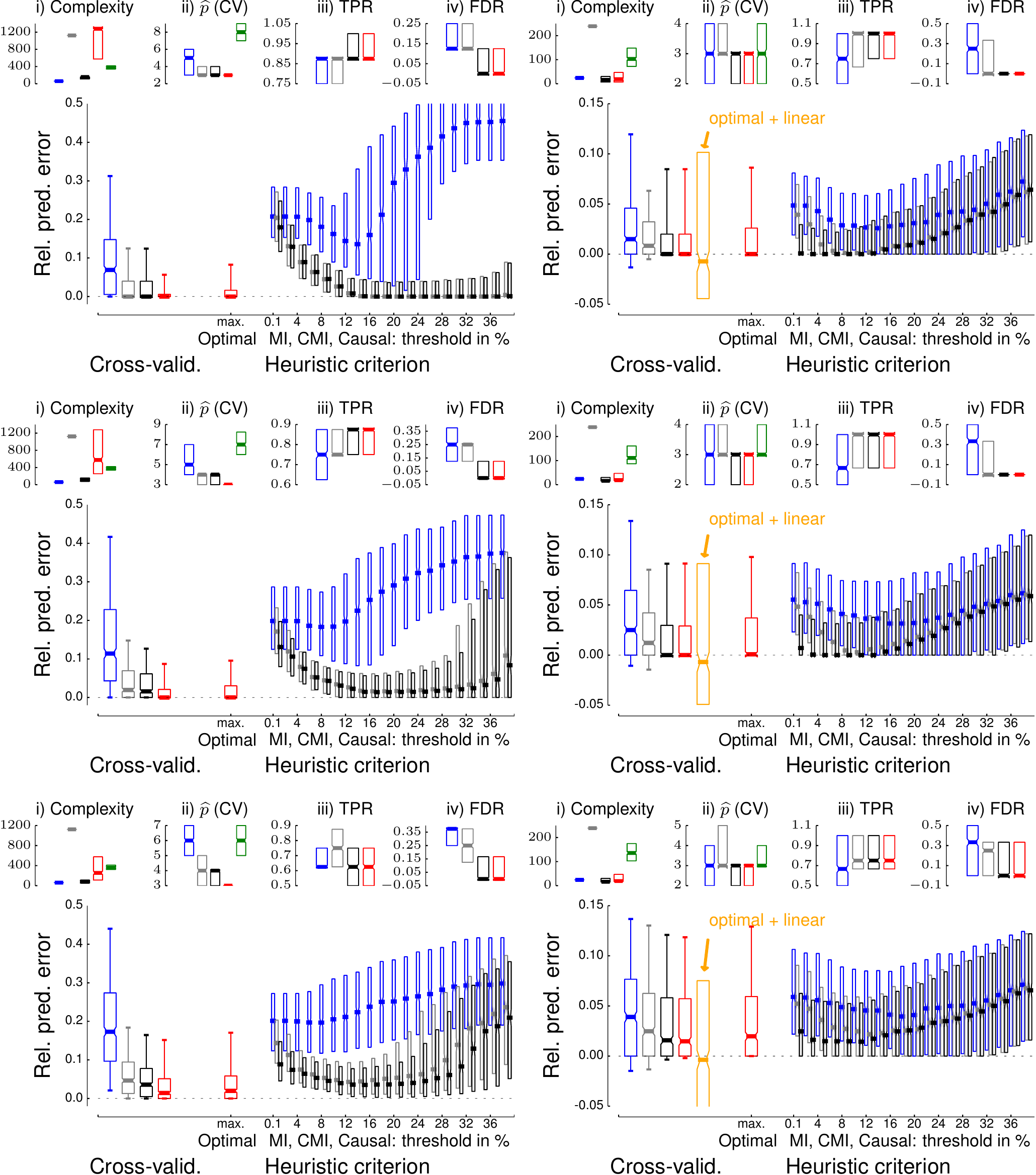}
\end{center}
\caption[]{(Color online) As in Fig.~\ref{fig:numerical_experiments} (left column) and Fig.~\ref{fig:prl_model} (right column) but for a larger range of time-series lengths $T=800,\,500,\,300$ (top to bottom, test set lengths $200,\,125,\,75$).}
\label{fig:robustness_length}
\end{figure*}
In Fig.~\ref{fig:robustness_sig} we show the results as in Fig.~\ref{fig:prl_model} for different values of the significance threshold $I^*$. Obviously this threshold affects the true positive and false discovery rate, which are, however, not directly of interest for the prediction task (as opposed to the causal inference problem). But a too low significance level in the causal pre-selection algorithm leads to a high computational complexity and also increases the variance in the optimal subset selection step, which results in higher prediction errors. If, on the other hand, the significance level is too high, too few predictors are available to optimize the prediction such that the resulting optimal predictors equal the pre-selected causal predictors $\mathcal{P}_{Y_{t+h}}$. If the significance level is adjusted to yield just a few predictors more than the number of optimal predictors $\widehat{p}$ (obtained through cross-validation or the optimal heuristic criterion), the prediction error is minimal and also the computational complexity is lower than for the non-causal CMI-forward selection scheme.

We also evaluate the prediction schemes for time series lengths $T=300$ and $T=800$. The results shown in Fig.~\ref{fig:robustness_length} demonstrate that the optimal scheme also works for very short time series and is even better for longer time series. For $T=800$ and the synergetic model~(\ref{eq:model_numerical_experiments}) the optimal scheme even results in 75\% of the realizations reaching the true minimal prediction error.

\bibliographystyle{unsrt} 

\end{document}